\pdfoutput=1

\documentclass[11pt]{article}

\usepackage{acl}

\usepackage{tikz}
\usepackage{times}
\usepackage{latexsym}

\usepackage[T1]{fontenc}
\usepackage[utf8]{inputenc}

\usepackage{microtype}
\usepackage{booktabs}

\usepackage{amssymb,amsmath}

\usepackage{graphicx}
\usepackage[caption=false]{subfig}
\DeclareMathOperator*{\argmax}{arg\,max}

\title{Combating the Curse of Multilinguality in Cross-Lingual WSD by Aligning Sparse Contextualized Word Representations}

\author{G\'abor Berend \\
  Institute of Informatics,
  University of Szeged \\
  \texttt{berendg@inf.u-szeged.hu}
}

\begin{document}
\maketitle
\begin{abstract}
	In this paper, we advocate for using large pre-trained monolingual language 
	models in cross lingual zero-shot word sense disambiguation (WSD) coupled 
	with a contextualized mapping mechanism. We also report rigorous 
	experiments that illustrate the effectiveness of employing sparse 
	contextualized word representations obtained via a dictionary learning 
	procedure. Our experimental results demonstrate that the above 
	modifications yield a significant improvement of nearly 6.5 points of 
	increase in the average F-score (from 62.0 to 68.5) over a collection of 17 
	typologically diverse set of target languages. We release our source code 
	for replicating our experiments at 
	\url{https://github.com/begab/sparsity_makes_sense}. 
\end{abstract}

\section{Introduction}

Word sense disambiguation (WSD) is a long-standing and fundamental problem of Natural Language Processing, known to be affected by the \emph{knowledge acquisition bottleneck} \citep{Gale1992}.
Large pre-trained neural language models are known to effectively mitigate the problems related to the paucity of high quality, large-coverage sense annotated training data for WSD (\citealp{loureiro-jorge-2019-language,10.1162/coli_a_00405}; \emph{inter alia}). 

Most recently, the knowledge acquisition bottleneck has been identified as an immense problem in the cross-lingual setting as well \cite{ijcai2020-687}. A straightforward solution for handling this problem is to apply large multilingual pre-trained language models in a zero-shot setting, however, this approach has a potential limitation owing to the \emph{curse of multilinguality} \cite{conneau-etal-2020-unsupervised}, i.e., the inability of such models to handle the large number of languages involved during training such models to an equally good quality.

The research community replied to the limitations of large massively multilingual models by developing language-specific monolingual language models.\footnote{With a slight abuse of notation, we also refer to models that support a handful of (related) languages (e.g. Slovenian and Croatian) as language-specific monolingual ones.} Table~\ref{tab:models} provides a shortlist of recently published monolingual large pre-trained language models, related to the languages involved in the cross-lingual WSD test suit, XL-WSD \cite{Pasini_Raganato_Navigli_2021}.

With the prevalence of large monolingual pre-trained models, the important research question arises if their language-specific nature can be successfully exploited during zero-shot learning.
Our research provides a thorough comparison of the application of large multilingual and monolingual pre-trained language models for zero-shot WSD.

\begin{table}
	\centering
	\resizebox{\columnwidth}{!}{
		\begin{tabular}{@{}rl@{}}
			\toprule
			ISO& Huggingface model identifier \\
			\midrule
			bg & \texttt{DeepPavlov/bert-base-bg-cs-pl-ru-cased}  \cite{arkhipov-etal-2019-tuning} \\
			ca & \texttt{PlanTL-GOB-ES/roberta-base-ca}  \cite{armengol-estape-etal-2021-multilingual} \\
			da & \texttt{Maltehb/danish-bert-botxo} \\
{de}
			 & \texttt{bert-base-german-cased} \\
		    es & \texttt{dccuchile/bert-base-spanish-wwm-cased} \cite{CaneteCFP2020} \\
            et & \texttt{EMBEDDIA/finest-BERT} \cite{ulcar-robnik2020finest}\\
			eu & \texttt{ixa-ehu/berteus-base-cased} \cite{agerri-etal-2020-give} \\
			fr & \texttt{camembert-base} \cite{martin2020camembert} \\
			gl & \texttt{dvilares/bertinho-gl-base-cased} \cite{DBLP:journals/pdln/VilaresGG21} \\
{hr} & \texttt{EMBEDDIA/crosloengual-bert}  \cite{ulcar-robnik2020finest} \\
			hu & \texttt{SZTAKI-HLT/hubert-base-cc} \cite{Nemeskey:2021a} \\
{it} & \texttt{Musixmatch/umberto-commoncrawl-cased-v1} \\
			ja & \texttt{cl-tohoku/bert-base-japanese-whole-word-masking}\\ 
			ko & \texttt{snunlp/KR-BERT-char16424} \\
			nl & \texttt{GroNLP/bert-base-dutch-cased} \cite{Vries2019BERTjeAD} \\
			sl & \texttt{EMBEDDIA/sloberta} \\
			zh & \texttt{bert-base-chinese} \\
			\bottomrule
		\end{tabular}
	}
	\caption{Monolingual models from the \texttt{transformers} library \cite{wolf-etal-2020-transformers} covering all the (non-English) languages of the XL-WSD dataset \citep{Pasini_Raganato_Navigli_2021}.}
\label{tab:models}
\end{table} 

Another crucial aspect that we carefully investigate in this paper is the integration of sparse contextualized word representations into cross-lingual zero-shot WSD.
Sparse word representations have a demonstrated ability to align with word senses \cite{Balogh_Berend_Diochnos_Turan_2020,yun-etal-2021-transformer}.
While the benefits of employing sparsity has been shown for WSD in English \cite{berend-2020-sparsity}, its viability in the cross-lingual setting has not yet been verified. In order to conduct such an analysis, we propose an algorithm for obtaining cross-lingual sparse contextualized word representations from independently trained monolingual language models.

\section{Related work}

The analysis and the investigation of the transfer capabilities of large pre-trained language models (such as mBERT or XLM) across languages has spurred significant research interest \cite{pires-etal-2019-multilingual,wu-dredze-2019-beto,wu-dredze-2020-languages,K2020Cross-Lingual}.
In contrast to the availability of multilingual neural language models, a series of recent papers have argued for the creation of dedicated neural language models for different languages (see e.g.~Table~\ref{tab:models}). 
While monolingual neural language models can more accurately model the distinct languages, models that are trained in isolation of other languages cannot directly benefit from downstream application-specific annotated training data available in different languages.

\citet{artetxe-etal-2020-cross} proposed an approach for making monolingual models compatible with each other by first pre-training a masked language model on a source language, then freezing its parameters apart from its embedding layer that get replaced and trained for additional target languages using a standard masked language modeling objective. Note that this approach is complementary and strictly more resource intensive to ours, as it involves the pre-training of a (freezed) transformer model with respect its embedding layer for a target language. In contrast, our approach can operate on monolingual language models fully pre-trained in total isolation from the source language encoder. Also, our approach learns substantially fewer parameters in the form of an alignment matrix between the hidden representations of the contextualized target and source language spaces. 

\citet{conneau-etal-2020-emerging} analyzed the multilingual patterns emerging in large pre-trained language models. The authors found that ``\emph{language universal representations emerge in pre-trained models without the requirement of any shared vocabulary or domain similarity}''. That work have demonstrated that monolingual BERT models can be effectively mapped for performing zero-shot cross-lingual named entity recognition and syntactic parsing. 
Similarly, \citet{wang-etal-2019-cross,schuster-etal-2019-cross} also illustrated the efficacy of linear transformations for using BERT-derived representations in cross-lingual dependency parsing.

WSD has been a fundamental and challenging problem in NLP for many decades, dating back to \cite{j:weaver}.
The utilization of contextualized word representations was first advocated by \citet{peters-etal-2018-deep}, later popularized by \citep{loureiro-jorge-2019-language,loureiro2021lmms}. \citet{ijcai2021-593} offers a survey of the recent approaches.

Most recently, \citet{rezaee-etal-2021-cross} have explored the usage of multilingual language models (XLM) in zero-shot WSD. While the experiments in \citep{rezaee-etal-2021-cross} cover four related target languages (German, Spanish, French and Italian), our investigation involves a typologically diverse set of 17 target languages (beyond English) from~\citep{Pasini_Raganato_Navigli_2021}. Our work also extends that line of research in important aspects, as we show that the application of monolingual neural language models can vastly improve the performance of cross-lingual zero-shot WSD. Additionally, we also provide a careful evaluation of sparse contextualized word representations in zero-shot WSD.

\citet{berend-2020-sparsity} introduced sparse contextualized word representations via the application of dictionary learning, and showed that  sense representations that are obtained from the co-occurrence statistics of the sparsity structure of the contextualized word representations and their sense annotations can provide significant improvement in monolingual WSD. Our work relates to that line of research by providing a mapping-based procedure, which enables the usage of such sense representations created in some source language to be applied in other target languages as well. The kind of mapping we employ can be viewed as a generalization of the approach introduced in \citep{Berend2020Massively} with the notable exception that in this work, we obtain sparse word representations for contextualized models as opposed to static word embeddings.

\section{Methodology}

In order to allow for zero-shot transfer between monolingual language models pre-trained in isolation from each other, we need to determine a mapping between their hidden representations. We first introduce our methodology for doing so, then we integrate this to the creation of sparse contextualized word representations.

\subsection{Mapping hidden representations}
\label{sec:mapping}

The alignment of word representations between independently constructed semantic spaces can be conveniently and efficiently performed via linear transformations. 
This has been a standard approach for non-contextualized word embeddings \cite{journals/corr/MikolovLS13,xing-etal-2015-normalized,ICLR2017-SmithTHH17}, but it has been shown to be useful in the contextualized case as well \citep{conneau-etal-2020-emerging}. 

The standard approach is to obtain a collection of pairs of anchor points $\{\boldsymbol{x}_i, \boldsymbol{y}_i\}_{i=1}^n$ with $\boldsymbol{x}_i$ and $\boldsymbol{y}_i$ denoting the representation of semantically equivalent words in the target and source languages, respectively. The mapping $W$ is then obtained as
\begin{equation}
	\min\limits_{W} \sum_{i=1}^{n}\lVert W\boldsymbol{x}_i -\boldsymbol{y}_i \rVert_2^2.
\end{equation}

As we deal with contextualized models, we can obtain various representations for a word even in the same context, by considering the hidden representations from different layers of the neural language models employed. Additionally, as constraining the mapping matrix to be an isometric one have proven to be a useful requirement, we define our learning task to be of the form 
\begin{equation}
	\min\limits_{W~s.t.~W^\intercal W = I} \sum_{i=1}^{n}\lVert W\boldsymbol{x}_i^{(l_t)} -\boldsymbol{y}_i^{(l_s)} \rVert_2^2, 
\end{equation}
with $I$ denoting the identity matrix, $\boldsymbol{x}_i^{(l_t)}$ and $\boldsymbol{y}_i^{(l_s)}$ denoting the hidden representations obtained from the $l_t$\textsuperscript{th} and $l_s$\textsuperscript{th} layers of the target and source language neural language models, respectively.

Finding the optimal isometric $W$ can be viewed as an instance of the 
orthogonal Procrustes problem \citep{Schonemann:1966ch} which can be solved by 
$W_\perp=UV$, with $U$ and $V$ originating from the singular value decomposition of the matrix product $Y^\intercal X$, where $X$ and $Y$ include the stacked target and source language contextual representations of pairs of semantically equivalent words.

As words of the input sequences to the neural language models can be split into multiple subtokens, we followed the common practice of obtaining word-level neural representations by performing mean pooling of the subword representations.
Throughout our experiments, we also relied on the RCSLS criterion \cite{joulin-etal-2018-loss}, which offers a retrieval-based alternative of obtaining a mapping from the target to the source language representations.

\subsection{Cross-lingual sparse contextualized word representations}

Our approach extends the information theoretic algorithm introduced in \cite{berend-2020-sparsity} for its application in the cross-lingual zero-shot WSD setting.
In order to obtain sparse contextualized representations for the source language, we first populate $Y \in \mathbb{R}^{d\times N}$ with $d$-dimensional contextualized representations of words determined for texts in the source language, and minimize the objective
\begin{equation}
	\min\limits_{D \in \mathcal{C}, \boldsymbol{\alpha_i} \in \mathbb{R}_{\geq0}^{k}} \sum_{i=1}^{N}\frac{1}{2} 
	\lVert \boldsymbol{y_i} - D\boldsymbol{\alpha_i} \rVert_2^2 + \lambda \lVert \boldsymbol{\alpha_i} \rVert_1,
	\label{eq:nonneg_SPAMS_objective}
\end{equation}
where $\mathcal{C}$ denotes the convex set of $d\times k$ matrices with column norm at most 1, $\lambda$ is a regularization coefficient and the sparse coefficients in $\alpha$ are required to be non-negative. 
We used the SPAMS library \citep{Mairal:2009:ODL:1553374.1553463} for 
calculating $D$ and~$\alpha$.

Having obtained $D$ for the source language, we determine a sparse 
contextualized word representation for a target language word with dense 
contextualized representation $\boldsymbol{x_i}$ as
\begin{equation}
	\min\limits_{\boldsymbol{\alpha_i} \in \mathbb{R}_{\geq0}^{k}} \frac{1}{2} 
	\lVert W\boldsymbol{x_i} - D\boldsymbol{\alpha_i} \rVert_2^2 + \lambda \lVert \boldsymbol{\alpha_i} \rVert_1,
	\label{eq:nonneg_SPAMS_objective_target}
\end{equation}
where $W$ is the alignment transformation as described earlier in Section~\ref{sec:mapping}. Eq.~\eqref{eq:nonneg_SPAMS_objective_target} reveals that the cross-lingual applicability of the sparse codes are assured by the mapping transformation $W$ and the fact that the sparse target language representations are also using the same $D$ that was determined for the source language, which also ensures the efficient calculation of sparse representations during inference time.

Apart from these crucial extensions we made for providing the use of contextualized sparse representations in the cross-lingual setting, the way 
we utilized them for the determination of sense representation and inference is identical to \cite{berend-2020-sparsity}. That is, for all sense-annotated words in the training corpus, we calculated a weighted co-occurrence statistics between a word pertaining to a specific semantic category and having non-zero coordinates along a specific dimension in their sparse contextualied word representations. These statistics are then transformed into pointwise mutual information (PMI) scores, resulting in a sense representation for all the senses in the training sense inventory.

Sense representations obtained that way measure the strength of the relation of the senses to the different (sparse) coordinates. Inference for a word with sparse representation $\boldsymbol{\alpha}$ is simply taken as $\argmax_{s}\Phi\boldsymbol{\alpha}^\intercal$, where $\Phi$ is the previously defined matrix of PMI values and $s$ corresponds to the sense at which position the above matrix--vector products takes its largest value.

\section{Experimental results}

All the neural language models that we relied on during our experiments were obtained from the \texttt{transformers} library \cite{wolf-etal-2020-transformers}. We used four NVIDIA Titan 2080 GPUs for our experiments.

As the multilingual language model, we used the 24-layer transformer architecture, XLM-RoBERTa (XLM-R for short) \cite{conneau-etal-2020-unsupervised}. We chose the cased BERT \cite{devlin-etal-2019-bert} large model as the monolingual model for encoding English text. As for the rest of the monolingual language models involved in our experiments, we relied on the models listed in Table~\ref{tab:models}. These monolingual models have the same size as the BERT-base model, i.e., they consist of 12 transformer blocks and employ hidden representations of 768 dimensions.

For evaluation purposes, we used the extra-large cross-lingual evaluation benchmark XL-WSD, recently proposed in \citep{Pasini_Raganato_Navigli_2021}.
The database contains a high-quality sense annotated corpus for English as the concatenation of the SemCor dataset \cite{miller-etal-1994-using} and the sense definitions and example sentences from WordNet \cite{wn}.
XL-WSD uses the unified cross-lingual sense inventory of BabelNet \cite{10.1016/j.artint.2012.07.001}.

The dataset contains 17 additional typologically diverse languages besides English (that we listed in Table~\ref{tab:models}). The authors also released machine translated silver standard sense annotated training corpora for all the languages, which makes the language-specific fine-tuning of monolingual models possible, however, as shown in \citep{Pasini_Raganato_Navigli_2021}, that approach resulted in inferior results compared to the application of multilingual models in the zero-shot setting.

Throughout the application of sparse contextualized representations, we employ the same set of hyperparameters that were used in \cite{berend-2020-sparsity}, i.e., we set the number of the regularization coefficient to $\lambda=0.05$ and the number of (sparse) coordinates to $k=3000$. There made one optional change, i.e., we decided whether to use the normalization of PMI values \cite{bouma2009normalized} during the calculation of the sense representation matrix $\Phi$ on a per language basis based on development set performances. An ablation study related to the (optional) normalization of PMI scores is reported in Table~\ref{tab:norm_ablation}, Appendix~\ref{sec:pmi}.

When we do not employ the sparsification of the contextualized word representations for determining the sense representations, we follow the approach introduced in \cite{loureiro-jorge-2019-language}. That is, we take the centroid of word vectors belonging to a particular sense as the representation of that sense, and perform a nearest neighbor search during inference.

\subsection{Alignment of contextualized representations}

As the different layers of neural language models have been shown to provide different levels of utility towards different tasks, we experimented with mappings between different combinations of layers from the target and source language neural language models. Since the last few layers of the neural models are generally agreed to be the most useful for semantics-related tasks \cite{peters-etal-2018-deep,tenney-etal-2019-bert,NEURIPS2019_159c1ffe}, we decided to learn mappings between the hidden representations of any of the last four layers of the target and source language encoders.

We used BERT as the language specific encoder for the source language texts in English, but we also investigated the application of XLM-R, so that we can see the effects of replacing it by an encoder especially tailored for English. As for the target languages, we used the respective models for each language as listed in Table~\ref{tab:models}.
Similar to the source language, we also investigated the case when target languages were encoded by the multilingual model.

In what follows, we label the different experimental settings according to the followings:
\begin{itemize}
	\item multi$\rightarrow$multi means that we map the target language representations obtained by the multilingual (XLM-R) model to the representation space of the source language also obtained by the multilingual (XLM-R) encoder,
	\item multi$\rightarrow$mono, means that we map the target language representations obtained by the multilingual (XLM-R) model to the representation space of the source language obtained by the monolingual (English BERT) encoder,
	\item mono$\rightarrow$multi, means that we map the target language representations obtained by their respective monolingual language model to the representation space of the source language obtained by the multilingual (XLM-R) encoder,
	\item mono$\rightarrow$mono, means that we map the target language representations obtained by their respective monolingual language model to the representation space of the source language obtained by the monolingual (English BERT) encoder.
\end{itemize}

In order to obtain the cross-representational mappings, we accessed the Tatoeba corpus \cite{tiedemann-2012-parallel} through the \texttt{datasets} library \cite{lhoest-etal-2021-datasets}. The Tatoeba corpus contains translated sentence pairs for several hundreds of languages which we used for obtaining the pivot word mention pairs together with their contexts. 

In addition to the Tatoeba corpus, we used the \texttt{word2word} library \cite{choe-etal-2020-word2word} containing word translation pairs between more than 3,500 language pairs. By denoting $(S_{s{_i}}, S_{t{_i}})$ the $i$\textsuperscript{th} translated sentence pair from the Tatoeba corpus, we treated those $(w_s \in S_{s{_i}}, w_t \in S_{t{_i}})$ word occurrences as being semantically equivalent, for which the $w_t \in TranslationOf(w_s)$ and the $w_s \in TranslationOf(w_t)$ relations simultaneously held according to the translation list provided by \texttt{word2word}.

As an example, given the German-English translation pair from Tatoeba, \emph{\{'de:' 'Es steht ein \underline{\textcolor{red}{Glas}} auf dem \underline{\textcolor{blue}{Tisch}}.', 'en': 'There is a \underline{\textcolor{red}{glass}} on the \underline{\textcolor{blue}{table}}.\}}, underlined pairs of words with the same color would be treated as contextualized translation pairs of each other. 

One benefit of our approach for determining contextual alignment of word pairs is that it does not require word level alignment of the parallel sentences, hence it suits such lower resource scenarios better, when only parallel sentences (without word level alignments) and a list of word translation pairs are provided.
Naturally, different contextual alignment approaches could be integrated into our approach at this point, and this is something that we regard as potential future extension of our work.

We evaluated the quality of the mapping learned between the target and the source language representations by defining a contextualized translation retrieval task and evaluating it on its accuracy@1 metric, i.e., for what fraction of the contextualized translation pairs -- not seen during the determination of the mapping between the two representation spaces -- are we able to rank the original translated context as the highest.

In the multi$\rightarrow$multi case, i.e., when both the target and source languages are encoded by the same multilingual model (XLM-R), it also makes sense to use the identity matrix as the mapping operator for mapping the target language contextual text representations to the semantic space of the source language (as long as the target and source language texts are obtained from the same layer of the multilingual encoder). We also evaluated the quality of this approach in our experiments that we refer to as the identity approach.

\begin{figure*}
	\centering
	\subfloat[Mapping accuracies per languages.\label{fig:top1}]{
		\includegraphics[width=\columnwidth]{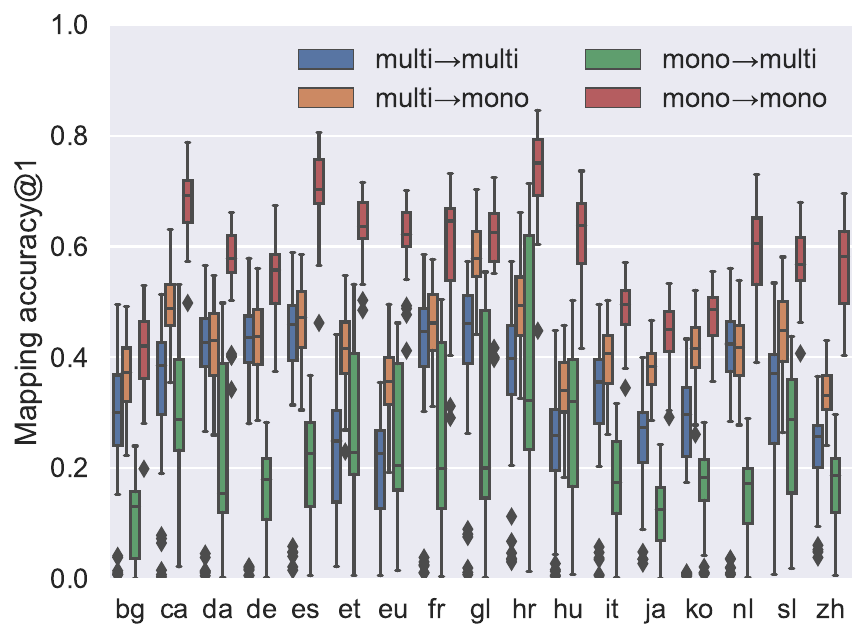}
	}
	\subfloat[Mapping accuracies aggregated over languages.\label{fig:top1_aggr}]{
		\includegraphics[width=\columnwidth]{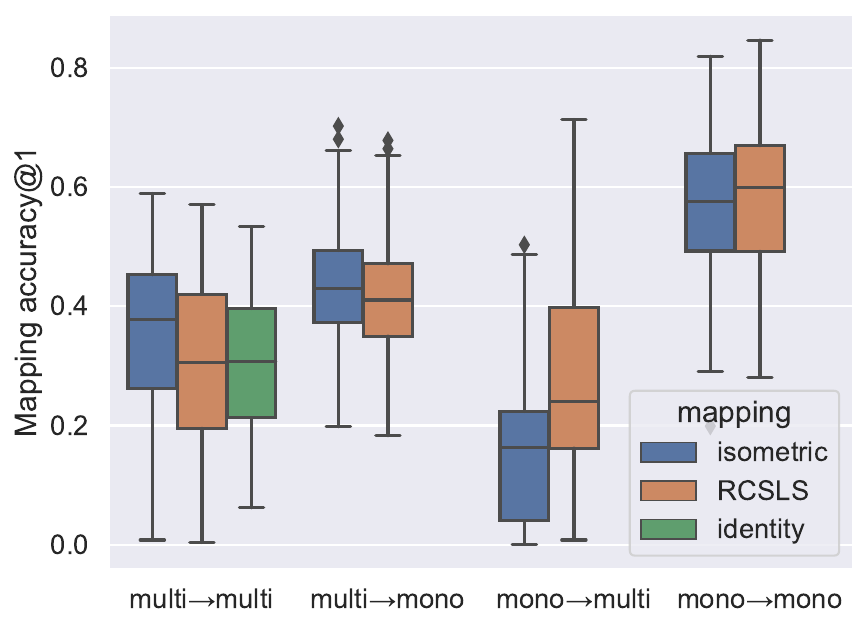}
	}
	\caption{The results of translation retrieval over the test sets of the different languages and different combinations of transformers used for the (English) source and the target languages.}
	\label{fig:mapping_acc}
\end{figure*}

We list the statistics of the Tatoeba corpus and the size of the training and test contextualized translation pairs in Table~\ref{tab:tatoeba}.
Our results on the top-1 contextualized translation retrieval accuracies along the different languages and combination of target and source encoder usage are reported in Figure~\ref{fig:mapping_acc}. The quality of the combination which uses monolingual encoders for both the target and source languages (mono$\rightarrow$mono) performed the best.

\begin{table}
	\centering
	\resizebox{\columnwidth}{!}{
	\begin{tabular}{llccc}
		\toprule
		\multicolumn{2}{l}{Language}& \#sentences & Train & Test \\ \midrule
		bg & Bulgarian & 17,797 & 14,212 & 3,554 \\
		ca & Catalan & 1,663 & 3,912 & 979 \\
		da & Danish & 30,089 & 20,000 & 5,000 \\
		de & German  & 299,769 & 20,000 & 5,000 \\
		es & Spanish & 207,517 & 20,000 & 5,000 \\
		et & Estonian & 2,428 & 2,365 & 592 \\
		eu & Basque & 2,062 & 3,956 & 990 \\
		fr & French & 262,078 & 20,000 & 5,000 \\
		gl & Galician & 1,013 & 2,356 & 590 \\
		hr & Croatian & 2,420 & 1,946 & 487 \\
		hu & Hungarian & 107,133 & 20,000 & 5,000 \\
		it & Italian & 482,948 & 20,000 & 5,000 \\
		ja & Japanese & 204,893 & 20,000 & 5,000 \\
		ko & Korean & 3,434 & 5,632 & 1,408 \\
		nl & Dutch & 72,391 & 20,000 & 5,000 \\
		sl & Slovenian & 3,210 & 1,285 & 322 \\
		zh & Chinese & 46,114 & 20,000 & 5,000 \\ \bottomrule
	\end{tabular}
}
\caption{The number of sentence pairs included in the Tatoeba corpus between English and a target language and the number of contextualized translation pairs extracted for training and testing the mappings.}
\label{tab:tatoeba}
\end{table}

\subsection{Monolingual evaluation}

We first conducted evaluations in the monolingual setting, i.e., we used the sense annotated training data to train and evaluate WSD models in English. 
The results of these experiments -- depending on the encoder architecture used (BERT/XLM-R), the layer of the encoder utilized (\{21,\ldots,24\}), and whether the sparsification of the contextualized representations took place (Dense/Sparse) -- are included in Table~\ref{tab:en_results}.

Unsurprisingly, the application of the language-specific BERT model achieved better scores compared to that of XLM-R. An interesting observation though, is that the drop in performance is much more subtle for those cases when the contextualized representations are enhanced via sparsification, i.e., the typical loss in performance across the layers is only 3 points (apart from the final layer), opposed to the typical loss of 4-7 points in the dense case. 

\begin{table}
	\centering
	\begin{tabular}{ccccc}
		\toprule
	& \multicolumn{2}{c}{BERT} & \multicolumn{2}{c}{XLM-R} \\ 
	Layer & Dense & Sparse & Dense & Sparse \\
	\cmidrule(r){2-3}  \cmidrule(r){4-5}
	21 & 74.39 & 77.45 & 69.29 & 74.51 \\
	22 & 74.87 & 77.60 & 67.87 & 74.50 \\
	23 & 74.45 & 77.86 & 67.48 & 74.26 \\
	24 & 73.58 & 76.21 & 64.50 & 70.06 \\
	\bottomrule
	\end{tabular}
\caption{English results expressed in F-score.}
\label{tab:en_results}
\end{table}

\subsection{Cross-lingual zero-shot evaluation}
\begin{table*}
	\centering
	\subfloat[Baseline results (MCS stands for Most Common Sense) from \cite{Pasini_Raganato_Navigli_2021}. \label{tab:baselines}]{
	\resizebox{\textwidth}{!}{
		\begin{tabular}{@{}lcccccccccccccccccp{1cm}@{}}
			\toprule
& bg & ca & da & de & es & et & eu & fr & gl & hr & hu & it & ja & ko & nl & sl & zh & Avg. \\ \midrule
	XLMR-Large&	\textbf{72.00}&	49.97&	80.61&	83.18&	75.85&	66.13&	47.15	&\textbf{83.88}& \textbf{66.28}& 72.29&	67.64&	\textbf{77.66}&	61.87&	64.20&	59.20& \textbf{68.36}&	51.62 & 65.66\\
	XLMR-Base&	71.59&	47.77&	79.18&	82.13&	76.55	&64.73&	43.86	&82.33&	64.79&	72.13&	68.36&	76.73&	61.46&	63.65&	58.77&	66.34&	49.77 &64.82\\
	MBERT&	68.78&	47.35&	76.04&	80.63&	74.66&	64.33&	42.41&	81.64&	68.07&	70.65&	65.24&	76.16&	60.34&	63.37&	56.64&	62.16&	48.99 &62.84\\
EWISER~(\citeyear{bevilacqua-navigli-2020-breaking}) & 68.64&	42.99&	76.67&	80.86&	71.85	&65.98	&42.85	&80.86	&59.41	&70.60	&66.17	&74.06	&55.77	&63.38	&57.50	&59.74	&48.30 & 62.16\\
	SyntagRank&	61.10	&43.98	&72.93	&75.99	&68.58	&56.31	&42.91	&69.57	&67.56	&68.35	&57.98	&69.57	&57.46	&50.29	&56.00	&52.25	& 41.23 & 57.68\\
	Babelfy& 60.39	&36.52	&71.33	&77.84	&64.07	&49.62	&36.65	&67.41	&64.17	&63.75	&51.99	&64.22	&51.91	&51.95	&44.27	&35.38 &34.94 & 52.85 \\
	MCS &58.16	&27.17	&64.33	&75.99	&55.65	&46.87	&32.72	&59.31	&60.85	&62.88	&47.29	&52.77	&48.71	&52.48	&44.61	&36.71	&29.62 & 49.13\\
\bottomrule
\end{tabular}
}
}
\\
	\subfloat[Our results relying on dense sense vectors.\label{tab:dense}]{
	\resizebox{\textwidth}{!}{
		\begin{tabular}{@{}lcccccccccccccccccp{1cm}@{}}
			\toprule
& bg & ca & da & de & es & et & eu & fr & gl & hr & hu & it & ja & ko & nl & sl & zh & Avg. \\ \midrule
multi & 67.07 & 47.46 & 76.58 & 80.74 & 70.61 & 65.23 & 42.53 & 75.60 & 56.85 & 70.63 & 65.42 & 71.38 & 58.45 & 63.88 & 54.86 & 61.91 & 48.98 & 61.98 \\
multi→multi & 68.99 & 51.62 & 78.56 & 80.51 & 70.02 & 65.28 & 44.68 & 78.62 & 57.44 & 71.59 & 68.99 & 70.90 & 61.40 & 64.41 & 57.73 & 61.17 & 50.65 & 63.71 \\
multi→mono & 68.82 & 44.17 & 79.75 & \textbf{84.69} & 70.88 & 64.68 & 40.95 & 79.66 & 56.58 & 71.34 & 68.07 & 69.93 & 59.71 & 64.49 & 59.25 & 61.37 & 50.77 & 63.30 \\
mono→multi & 69.68 & 52.95 & 78.90 & 82.02 & 68.34 & 66.33 & 49.62 & 80.17 & 58.30 & 72.34 & 70.75 & 74.01 & 64.35 & 65.02 & 59.32 & 64.76 & 54.95 & 65.57 \\
mono→mono & 71.17 & 53.31 & 81.21 & 83.29 & 72.56 & 66.48 & 51.08 & 81.55 & 63.14 & 73.76 & 72.76 & 72.52 & 65.26 & 66.57 & 60.52 & 67.42 & 55.45 & 66.96 \\
\bottomrule
	\end{tabular}
}
}
\\
	\subfloat[Our results based on sparse sense vectors.\label{tab:sparse}]{
	\resizebox{\textwidth}{!}{
		\begin{tabular}{@{}lcccccccccccccccccp{1cm}@{}}
\toprule
& bg & ca & da & de & es & et & eu & fr & gl & hr & hu & it & ja & ko & nl & sl & zh & Avg. \\ \midrule
multi & 70.69 & 51.52 & \textbf{81.41} & 83.53 & 76.45 & 67.78 & 47.85 & 83.62 & 64.47 & 73.06 & 70.10 & 76.65 & 63.73 & 64.67 & 58.00 & 64.12 & 53.29 & 66.04 \\
multi→multi & 70.91 & 51.52 & 80.50 & 82.37 & 75.96 & 66.13 & 47.09 & 83.79 & 63.26 & 72.94 & 70.01 & 77.17 & 64.47 & 64.73 & 60.16 & 66.49 & 53.05 & 66.15 \\
multi→mono & 71.91 & 50.54 & 81.21 & 79.93 & \textbf{76.93} & 64.83 & 44.05 & 83.62 & 62.87 & 71.64 & 69.26 & 77.48 & 63.59 & 64.59 & 60.39 & 61.07 & 53.48 & 65.82 \\
mono→multi & 70.76 & 52.49 & 79.67 & 82.25 & 75.09 & 67.83 & 50.89 & 83.19 & 60.68 & 73.99 & 72.97 & 75.33 & 63.80 & 65.86 & 61.57 & 65.70 & 55.65 & 66.79 \\
mono→mono & \textbf{72.00} & \textbf{57.47} & 81.15 & 83.76 & 76.12 & \textbf{68.88} & \textbf{51.71} & 83.10 & 63.92 & \textbf{74.40} & \textbf{75.52} & 76.12 & \textbf{67.47} & \textbf{67.52} & \textbf{61.95} & 67.47 & \textbf{57.05} & \textbf{68.47} \\
\bottomrule
	\end{tabular}
	}
}
\caption{Test set results on the XL-WSD benchmark. The hyperparameters of the individual approaches (e.g. which layer of the target language encoder to align with which layer of the source language encode) were determined based on the development set of each language.}
\label{tab:model_selected_results}
\end{table*}

Table~\ref{tab:model_selected_results} includes the zero-shot cross-lingual WSD results for a collection of baseline approaches (Table~\ref{tab:baselines}) from~\cite{Pasini_Raganato_Navigli_2021}, followed by our models not utilizing the sparsification of the contextualized embeddings (Table~\ref{tab:dense}) and the ones that additionally benefit from sparsification as well (Table~\ref{tab:sparse}). It is useful to note that the mono→* approaches are strictly more resource efficient during inference as they are based on 12-layer encoders instead of the 24 layers of the multilingual XLM-R model.

At this point, we separate the multi→multi results into two, i.e., 1) those obtained when relying on the hidden representations from the same layer of XLM-R without mapping (or equivalently, with the identity mapping from the target to source representations); and 2) those obtained when the target and source language contextual representations could originate from different layers of the XLM-R encoder, and a non-identity (either isometric or RCSLS) mapping was employed. We keep referring to the latter as multi→multi, and denote the former type of experiments as multi (without the →multi suffix as there were no real mappings performed in these cases). 
Inspecting the first two rows of Table~\ref{tab:dense} and Table~\ref{tab:sparse} reveals that enhancing the multilingual encoder towards the treatment of a particular pair of languages by providing it a language pair specific mapping has a larger positive effect when using dense vectors. In fact, it increased the micro-averaged F-score over the 17 languages by 1.72 and 0.11 points for the dense and the sparse cases, respectively.

Overall, the micro-averaged F-score of our final approach managed to improve nearly 6.5 points (cf.~the first row of Table~\ref{tab:dense} and the last row in Table~\ref{tab:sparse}). A 5 point average improvement is due to the replacement of the XLM-R encoder for both the source language during training and target languages for inference (cf.~the first and last row of Table~\ref{tab:dense}) and an additional 1.5 points of improvement was an effect of our sparsification in the cross-lingual setting. The inspection of the third and fourth rows in both Table~\ref{tab:dense} and Table~\ref{tab:sparse} reveals that using a monolingual encoder during inference helps more compared to the application of a monolingual encoder for encoding the source language during training. 

We conducted the McNemar test between our system outputs when a non-identity mapping was used between a pair of languages. Our investigation revealed that all such $\binom{8}{2}$ pairs of system outputs from Table~\ref{tab:dense} and Table~\ref{tab:sparse} differ significantly from each other with $p<0.0007$, with only four exceptions, i.e, 1) multi→multi and multi→mono from Table~\ref{tab:dense}; 2) multi→multi and multi→mono from Table~\ref{tab:sparse}; 3) mono→multi from Table~\ref{tab:sparse} and mono→mono from Table~\ref{tab:dense}; 4) multi→mono from Table~\ref{tab:sparse} and mono→multi from Table~\ref{tab:dense}. 

\begin{figure}
	\centering
	\includegraphics[width=.94\linewidth]{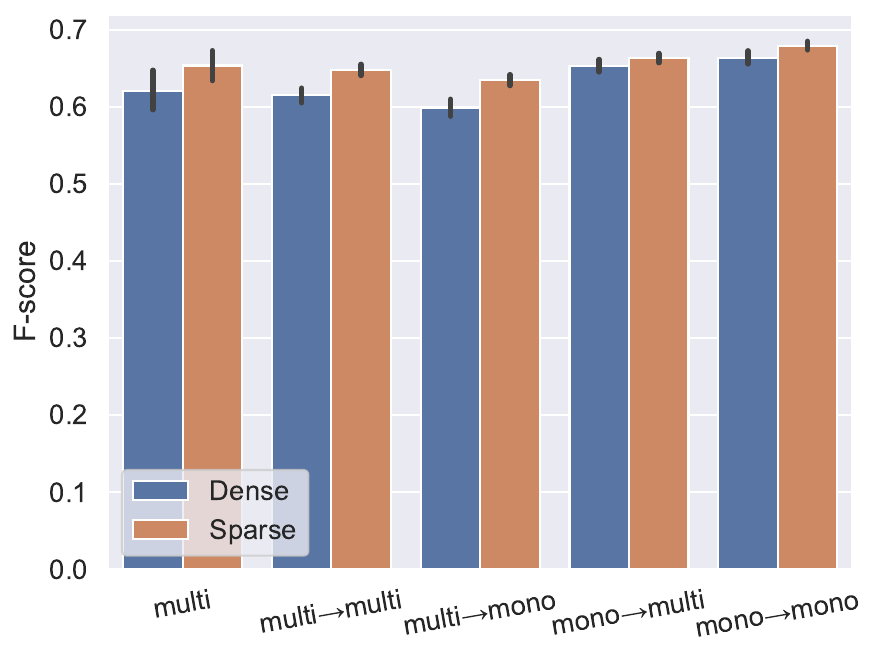}
	\caption{Overall averaged results for all the experiments conducted for the different approaches.}
	\label{fig:overall_res}
\end{figure}

Figure~\ref{fig:overall_res} summarizes the results of all the possible runs conducted. When using the multilingual encoder for both the target and source languages without a mapping step between the two (multi), we ran 4 different experiments per each language based on the hidden representations obtained from one of the last 4 layers of the multilingual encoder.
For the remaining experiments relying on the dense and sparse representations, there were 32 and 64 experiments for each language, respectively.
The 32 experiments were a result of choosing any of the 16 possible combination of the final four layers on the target and source language encoder, coupled with the type of mapping utilized (isometric/RCSLS). For the experiments involving the sparse representations, there was an extra parameter, whether the normalization of the PMI scores for obtaining the sense representations to be performed, resulting in $2 \times 32$ experiments all together. Our ablation study in Table~\ref{tab:norm_ablation} illustrates that this extra factor of $2$ for the sparse experiments did not provided us an unfair advantage, i.e., when fixing the value of normalization in any way, the overall results did not differ substantially.

The difference in the average performance of our approach transforming sparse contextualized representations obtained by monolingual models is significant (using unpaired t-test\footnote{We used unpaired t-test as the number of experiments was not same in all cases, i.e., 4 experiments/language in the multi case, and either 32 or 64 experiments/language in the rest of the cases.}, $p<0.005$) compared to any other configuration. This suggests that the mono$\rightarrow$mono approach has a robust advantage over alternative variants, and the improvements seen in Table~\ref{tab:model_selected_results} are \emph{not} an effect of careful hyperparameter selection, but they generalize over a wide range of choices.

\begin{figure}
	\centering
	\includegraphics[width=\linewidth]{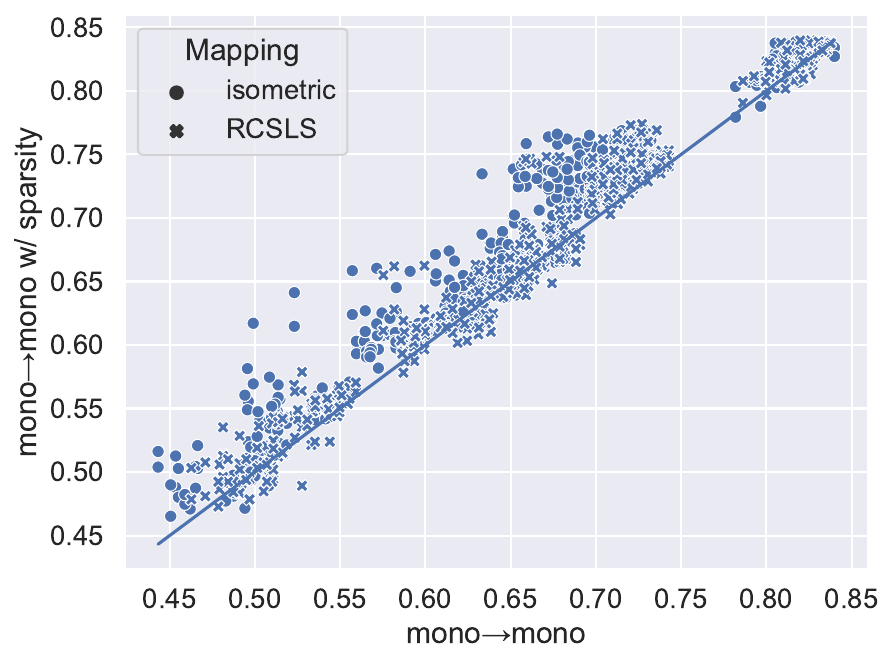}
	\caption{Comparison of the two best performing systems when the same hyperparameters were employed.}
	\label{fig:comparison}
\end{figure}

This effect is further corroborated in Figure~\ref{fig:comparison}, which offers a comparison between the two systems with the best average performance, i.e., mono$\rightarrow$mono that operates with the dense vectors (results are along the x-axis) and the same model but with the enhancement of sparsification (results are along the y-axis). Each data point corresponds to a setting with the same hyperparameter choices, and points above the diagonal line with slope one demonstrate the benefits of sparsification.

We have demonstrated the improved utility of mapping language-specific sparse contextualized representations for conducting zero-shot WSD, requiring large pre-trained language-specific text encoders for the target languages. While such models are available for all languages in  XL-WSD, a variety of the existing languages lack their dedicated language-specific pre-trained language model.

As such, an important question emerges whether it is possible to enjoy the benefits of mapping sparse contextualized representations for zero-shot WSD in the absence of a large pre-trained language model dedicated to the target language. 
To this end, we shall inspect the results of our multi→mono approach in Table~\ref{tab:model_selected_results}, a series of mapping-based experiments in which we acted as if the monolingual language models (other than the one for English) did not exist. In these experiments, the sense embeddings were obtained with bert-large-cased (being specialized to English), and the mapping to the non-English target languages were performed towards their XLM-R representations during the evaluation. This way, we could simulate the effects of the absence of language-specific models.

The multi→mono approach provided a substantially better average performance compared to the mere utilization of a multilingual encoder in the case of dense contextualized representations as it can be seen in Table~\ref{tab:dense}.
The average results of multi→mono are slightly inferior (albeit statistically insignificantly) to that of the multi approach for the application of sparse contextualized representations. 
However, when comparing the multi→multi results with that of multi→mono, we can see that by relying on a multilingual encoder alone, and allowing a mapping to be employed between its hidden representations pertaining to different languages, one can obtain the same (or even slightly better) performance as with the multi→mono approach. This highlights the importance of monolingual encoders for the target language, which seems to be more important than having access to a monolingual encoder for the source language.

\section{Conclusions}

In this paper we provided a systematic investigation of the benefits of using large monolingual pre-trained language models in place of multilingual language models, such as XLM-R. We have shown that since monolingual neural language models are specifically tailored for a single (or at most a few related) languages, they can effectively mitigate the \emph{curse of multilinguality} typical of multilingual models, and their application can significantly improve the F-scores in zero-shot WSD.
We additionally showed that the benefits of sparse contextualized word representations, obtained via a dictionary learning procedure, also convey to the cross-lingual setting, and that it provides complementary improvements to the usage of monolingual neural language models.

\section*{Acknowledgments}
The research was supported by the Ministry of Innovation and
Technology NRDI Office within the framework of the Artificial
Intelligence National Laboratory Program.
Additionally, we are thankful for the usage of ELKH Cloud
(https://science-cloud.hu/) that helped us achieving the results
published in this paper.

\bibliography{paper}
\bibliographystyle{acl_natbib}

\appendix 

\section{Analysis of the language models}
\label{sec:appendix}

\begin{figure*}[!ht]
	\centering
	\subfloat[Fertility\label{fig:models_fertility}]{
		\includegraphics[width=.9\columnwidth]{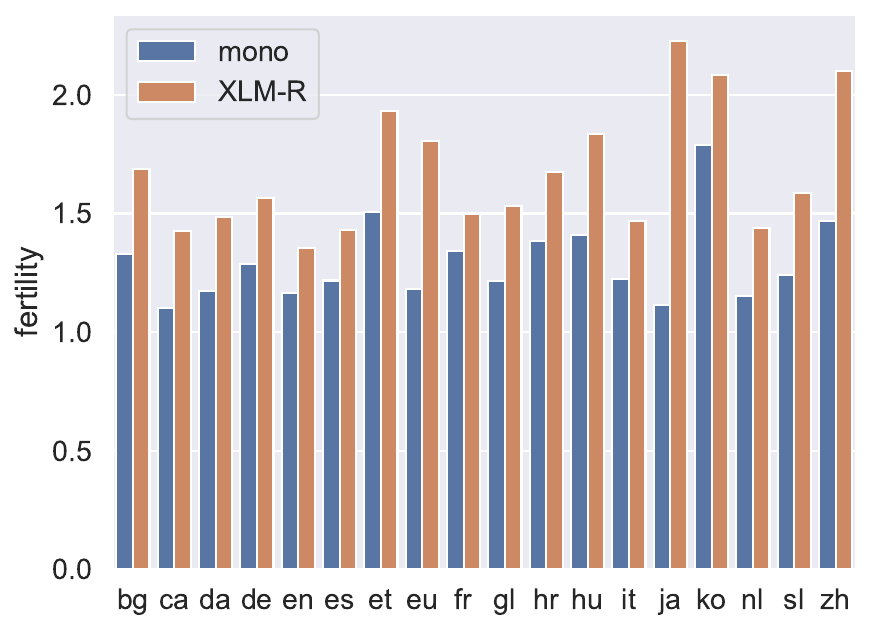}
	}~
	\subfloat[Multi-token ratio (MTR)\label{fig:models_mtr}]{
		\includegraphics[width=.9\columnwidth]{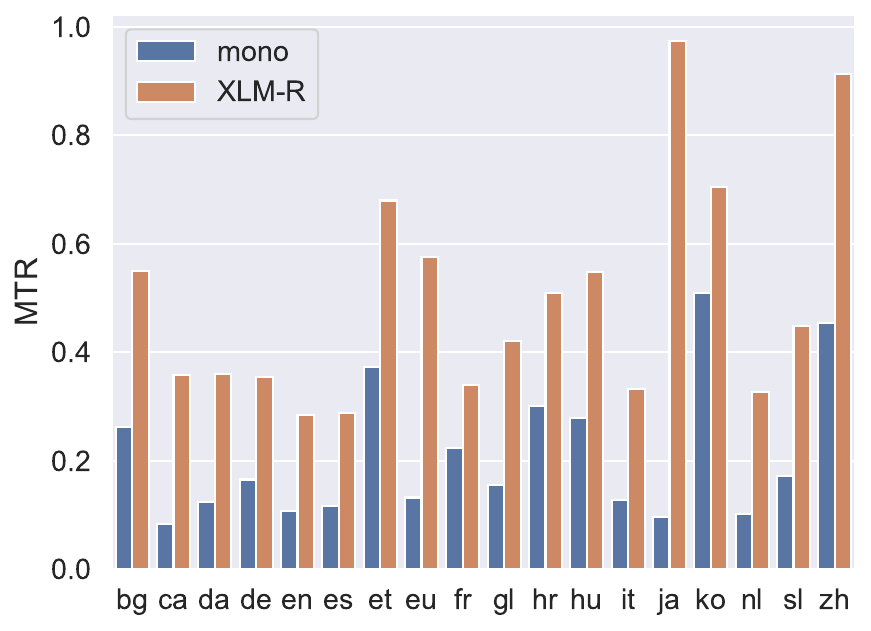}
	}
	\caption{Comparison of the tokenizers of the multilingual (XLM-R) and the monolingual language models.}
	\label{fig:models_tok}
\end{figure*}

\begin{figure*}
	\centering	
	\subfloat[MLM loss of the monolingual models\label{fig:models_MLM_mono}]{
		\includegraphics[width=.9\columnwidth]{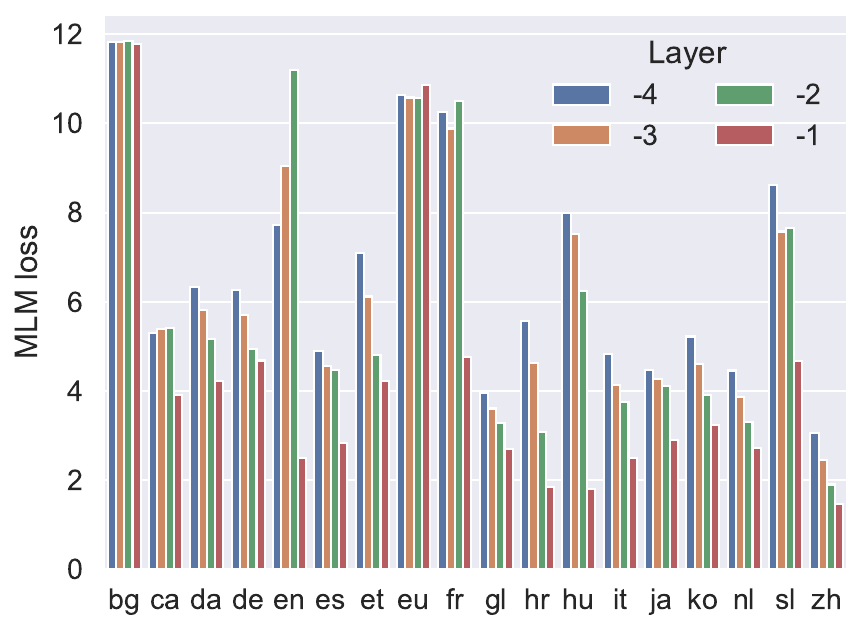}
	}
	\subfloat[MLM loss of XLM-R\label{fig:models_MLM_multi}]{
		\includegraphics[width=.9\columnwidth]{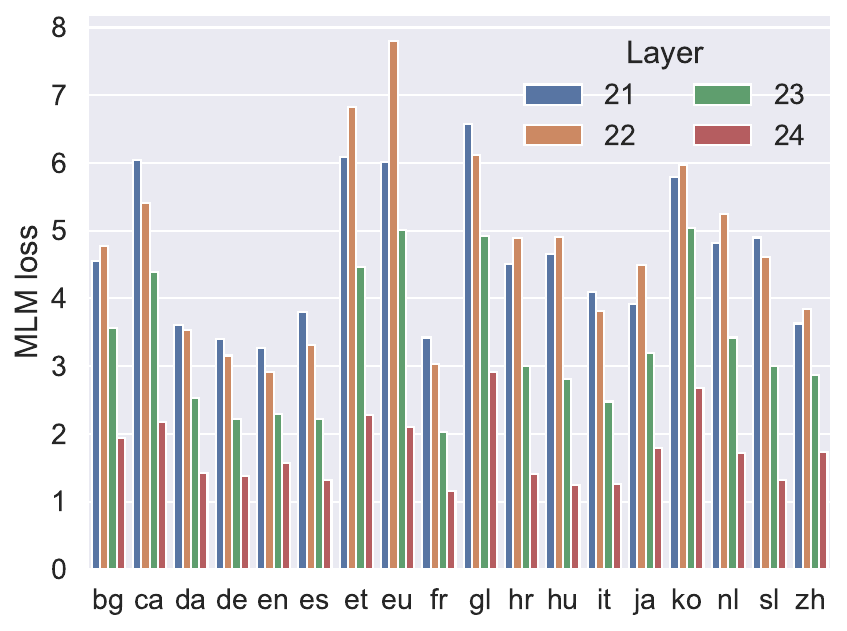}
	}
	\caption{The comparison of the multilingual (XLM-R) and the monolingual neural language models in terms of subword tokenization and their MLM objective.}
	\label{fig:models_mlm}
\end{figure*}

We compare some of the basic properties of the pretrained language models that we employed in Figure~\ref{fig:models_tok} and Figure~\ref{fig:models_mlm}. This can be useful as the monolingual quality of the language models we used could influence and account for their utility when used in conjunction with our mapping-based algorithm.
 
Figure~\ref{fig:models_tok} includes quantitative scores over the different languages related to the subword tokenizers employed by the various language models.
Fertility in Figure~\ref{fig:models_fertility} refers to the average number of subtokens a single token gets separated into by the tokenizer of the given language model. Multi-token ratio (MTR) in Figure~\ref{fig:models_mtr} indicates the fraction of tokens that gets split into more than one piece upon tokenization \cite{acs:2019,rust-etal-2021-good}. Smaller values of MTR mean a better adaptation of the tokenizer to the peculiarities of the given language. It can be seen that the monolingual models do a much better job compared to XLM-R, which can be part of the reason why mapping independently trained monolingual .

In Figure~\ref{fig:models_MLM_mono}, we refer to the last four layers of the investigated models as \{-4,-3,-2,-1\} as the English BERT is a 24-layer model, whereas the rest of the monolingual models consist of 12 layers. 
This means that layer -1 refers to layer 24 for English and layer 12 for some non-English model. Even though Figure~\ref{fig:models_MLM_mono} shows pathological masked language modeling (MLM) losses for certain monolingual models (e.g. Bulgarian or Basque) when measured on the XL-WSD database, their mapping-based utilization in zero-shot WSD was still possible as indicated by our main results (see Table~\ref{tab:model_selected_results}). A further interesting phenomenon is that the performance of XLM-R exceeds that of the bert-large-cased model in terms of MLM for English. 
These results suggest that the masked language modeling performance of pretrained language models and their utility in WSD are not strongly related with each other.

\section{Analysis on using the normalization of PMI scores}
\label{sec:pmi}
Upon the calculation of the sense representation matrix $\Phi$, involving the calculation of PMI scores between the various senses from the sense inventory and the coordinates of a sparse contextual representation being non-zero, \citet{berend-2020-sparsity} suggested the use of normalized PMI scores \citep{bouma2009normalized}.
Our preliminary results suggested that the normalization of PMI scores can have a mixed effect over the different languages. Table~\ref{tab:norm_ablation} includes a detailed breakdown on this effect for the individual languages.

\begin{table*}[!ht]
	\subfloat[Our results based on sparse sense vectors when \emph{always} using the normalization of PMI scores as done in \cite{berend-2020-sparsity}.\label{tab:norm}]{
		\resizebox{\textwidth}{!}{
			\begin{tabular}{@{}lcccccccccccccccccp{1cm}@{}}
				\toprule
				multi & 70.69 & 49.46 & 81.38 & 83.53 & 76.45 & 67.23 & 47.22 & 83.62 & 64.47 & 72.16 & 68.29 & 76.65 & 62.79 & 64.70 & 59.02 & 67.96 & 53.29 & 65.82 \\
				multi→multi & 70.91 & 49.31 & 80.81 & 82.37 & 75.96 & 67.28 & 44.30 & 83.79 & 62.63 & 71.89 & 69.24 & 77.17 & 63.26 & 64.52 & 60.16 & 66.49 & 52.82 & 65.71 \\
				multi→mono & 71.91 & 48.79 & 81.64 & 79.93 & 76.93 & 64.83 & 42.72 & 83.62 & 62.87 & 71.34 & 69.29 & 77.48 & 62.58 & 64.59 & 60.68 & 63.09 & 52.44 & 65.54 \\
				mono→multi & 70.76 & 50.49 & 79.93 & 83.41 & 75.09 & 66.13 & 49.37 & 83.19 & 60.68 & 73.11 & 71.66 & 75.33 & 63.01 & 64.44 & 60.70 & 66.63 & 55.14 & 66.23 \\
				mono→mono & 72.00 & 54.90 & 81.27 & 83.76 & 76.12 & 67.28 & 49.87 & 83.10 & 63.92 & 73.33 & 74.12 & 76.12 & 65.57 & 66.31 & 61.55 & 67.47 & 56.32 & 67.69 \\
				\bottomrule
			\end{tabular}
		}
	}
	
	\subfloat[Our results based on sparse sense vectors when \emph{not} using the normalization of PMI scores as done in \cite{berend-2020-sparsity}.\label{tab:no_norm}]{
		\resizebox{\textwidth}{!}{
			\begin{tabular}{@{}lcccccccccccccccccp{1cm}@{}}
				\toprule
				& bg & ca & da & de & es & et & eu & fr & gl & hr & hu & it & ja & ko & nl & sl & zh & Avg. \\ \midrule
				multi & 69.62 & 51.52 & 81.41 & 82.25 & 74.01 & 67.78 & 47.85 & 82.76 & 62.79 & 73.06 & 70.10 & 74.32 & 63.73 & 64.67 & 58.00 & 64.12 & 53.62 & 65.68 \\
				multi→multi & 69.97 & 51.52 & 80.50 & 82.13 & 74.07 & 66.13 & 47.09 & 82.76 & 63.26 & 72.94 & 70.01 & 74.63 & 64.47 & 64.73 & 60.16 & 65.55 & 53.05 & 65.82 \\
				multi→mono & 71.56 & 50.54 & 81.21 & 83.18 & 74.45 & 65.68 & 44.05 & 79.05 & 61.77 & 71.64 & 69.26 & 74.19 & 63.59 & 64.41 & 60.39 & 61.07 & 53.48 & 65.51 \\
				mono→multi & 70.16 & 52.49 & 79.67 & 82.25 & 70.77 & 67.83 & 50.89 & 81.29 & 58.65 & 73.99 & 72.97 & 73.92 & 63.80 & 65.86 & 61.57 & 65.70 & 55.65 & 66.42 \\
				mono→mono & 71.31 & 57.47 & 81.15 & 82.25 & 72.29 & 68.88 & 51.71 & 81.38 & 61.03 & 74.40 & 75.52 & 73.49 & 67.47 & 67.52 & 61.95 & 65.94 & 57.05 & 67.96 \\
				\bottomrule
			\end{tabular}
		}
	}
	
	\subfloat[Our results based on sparse sense vectors when the normalization of PMI scores was \emph{optional} and based on the development set for each language.\label{tab:optional_norm}]{
		\resizebox{\textwidth}{!}{
			\begin{tabular}{@{}lcccccccccccccccccp{1cm}@{}}
				\toprule
				& bg & ca & da & de & es & et & eu & fr & gl & hr & hu & it & ja & ko & nl & sl & zh & Avg. \\ \midrule
				multi & 70.69 & 51.52 & 81.41 & 83.53 & 76.45 & 67.78 & 47.85 & 83.62 & 64.47 & 73.06 & 70.10 & 76.65 & 63.73 & 64.67 & 58.00 & 64.12 & 53.29 & 66.04 \\
				multi→multi & 70.91 & 51.52 & 80.50 & 82.37 & 75.96 & 66.13 & 47.09 & 83.79 & 63.26 & 72.94 & 70.01 & 77.17 & 64.47 & 64.73 & 60.16 & 66.49 & 53.05 & 66.15 \\
				multi→mono & 71.91 & 50.54 & 81.21 & 79.93 & 76.93 & 64.83 & 44.05 & 83.62 & 62.87 & 71.64 & 69.26 & 77.48 & 63.59 & 64.59 & 60.39 & 61.07 & 53.48 & 65.82 \\
				mono→multi & 70.76 & 52.49 & 79.67 & 82.25 & 75.09 & 67.83 & 50.89 & 83.19 & 60.68 & 73.99 & 72.97 & 75.33 & 63.80 & 65.86 & 61.57 & 65.70 & 55.65 & 66.79 \\
				mono→mono & 72.00 & 57.47 & 81.15 & 83.76 & 76.12 & 68.88 & 51.71 & 83.10 & 63.92 & 74.40 & 75.52 & 76.12 & 67.47 & 67.52 & 61.95 & 67.47 & 57.05 & 68.47 \\
				\bottomrule
			\end{tabular}
		}
	}
	\caption{The effects of making the normalization of PMI scores \cite{bouma2009normalized} (a)~mandatory, (b)~prohibited, (c)~optional to use (based on development set results) during the creation of the sparse sense representations.}\label{tab:norm_ablation}
\end{table*}

\end{document}